\documentclass{article}
\usepackage{ijcai22}

\usepackage{times}
\usepackage{soul}
\usepackage{url}
\usepackage[hidelinks]{hyperref}
\usepackage[utf8]{inputenc}
\usepackage[small]{caption}
\usepackage{graphicx}
\usepackage{amsmath}
\usepackage{amsthm}
\usepackage{booktabs}
\usepackage{algorithm}
\usepackage{algorithmic}
\usepackage{color,xcolor}
\usepackage{amssymb}
\usepackage{float}
\usepackage{subfig}
\usepackage{color}
\DeclareMathOperator*{\argmax}{argmax}
\DeclareMathOperator*{\argmin}{argmin}

\title{Conservative Distributional Reinforcement Learning with Safety Constraints \footnote{This work has been submitted to the IEEE for possible publication. Copyright may be transferred without notice, after which this version may no longer be accessible.}}

\author{
    Hengrui Zhang, Youfang Lin, Sheng Han, Shuo Wang, Kai Lv 
    \affiliations
    Beijing Jiaotong University
    \emails
    \{18112037,yflin,shhan,shuo.wang,lvkai\}@bjtu.edu.cn
}

\begin{document}
\textbf{This work has been submitted to the IEEE for possible publication. Copyright may be transferred without notice, after which this version may no longer be accessible.}
\maketitle

\begin{abstract}
Safety exploration can be regarded as a constrained Markov decision problem where the expected long-term cost is constrained. 
Previous off-policy algorithms convert the constrained optimization problem into the corresponding unconstrained dual problem by introducing the Lagrangian relaxation technique. 
However, the cost function of the above algorithms provides inaccurate estimations and causes the instability of the Lagrange multiplier learning.
In this paper, we present a novel off-policy reinforcement learning algorithm called Conservative Distributional Maximum a Posteriori Policy Optimization (CDMPO). 
At first, to accurately judge whether the current situation satisfies the constraints, CDMPO adapts distributional reinforcement learning method to estimate the Q-function and C-function. 
Then, CDMPO uses a conservative value function loss to reduce the number of violations of constraints during the exploration process.
In addition, we utilize Weighted Average Proportional Integral Derivative (WAPID) to update the Lagrange multiplier stably. 
Empirical results show that the proposed method has fewer violations of constraints in the early exploration process.
The final test results also illustrate that our method has better risk control.
\end{abstract}

\section{Introduction}
Safety exploration \cite{garcia2015comprehensive,bharadhwaj2020conservative,yang2021wcsac} is a crucial issue for applying reinforcement learning to practical problems. Traditional reinforcement learning uses trial and error to interact with the environment to maximize returns. 
However, this is often not allowed in many safety exploration scenarios. 
For example, in expensive robotic platforms, researchers focus not only on maximizing expected returns, but also on preventing hardware damage \cite{garcia2015comprehensive}.

A natural way to consider safety in reinforcement learning is to introduce safety constraints. Thus, the traditional Markov Decision Process (MDP) is generalized to the Constrained Markov Decision Process (CMDP). 
Some existing attempts utilize off-policy algorithms to perform CMDP.
However, the constraint effect has not achieved excellent results.
The main problems are as follows:
1) the value estimation of the algorithm is inaccurate; 
2) the exploration process of the algorithm is not safe.

For the first problem, we analyse its reasons and adopt a \textit{distributional} approach to accurately estimate the Q-function and C-function. 
Inaccurate estimation exists in almost all reinforcement learning approaches. 
Moreover, off-policy methods would introduce additional estimation biases and variances as the behavior policy is not the evaluation policy. 
Thus, using off-policy data to estimate the C-function is inaccurate. 
Furthermore, inaccurate C-function may cause the agent to make a wrong judgment on whether the constraints are satisfied in the current situation, leading to unsafe actions. 
Existing distributional RL algorithms \cite{bellemare2017distributional,duan2021distributional} utilize the return distribution to adaptively adjust the update step size of Q-values.
Inspired by the above works, we introduce the distributional approach to evaluate the Q-function and the C-function.

For the second problem, we propose a \textit{conservative} method that consists of a \textit{conservative exploration strategy} and a \textit{conservative C-function update method}. 
Existing arts perform exploration with the action that is sampled from the distribution produced by the actor network. 
Note that the above action is easier to violate constraints, thus making the exploration process unsafe. 
In this work, we propose a conservative exploration strategy that filters actions by the C-function to obtain the most conservative action. 
The conservative C-function update method consists of a Temporal Difference (TD) loss and an additional regularization term. Specifically, the regularization term guides the C-function to output a lower risk for the conservative action than the rest actions. 

In this paper, we utilize Lagrange Multiplier to introduce safety constraints.
However, utilizing gradient descent methods to optimize the Lagrange multiplier is unstable in practice. 
Thus, we first attempt to adopt Proportional Integral Derivative (PID) to damp oscillations, but the problem still exists. 
We then propose Weighted Average PID (WAPID) to solve the oscillation problem. 

In summary, we propose CDMPO to solve the safety exploration task, and it has the following contributions.
\begin{itemize}
\item We adopt distributional reinforcement learning method to estimate the Q-function and C-function, which can accurately judge whether the current situation satisfies the constraints.
\item We propose a conservative exploration strategy and a conservative C-function update method to make the agent perform exploration safely. 
\item We utilize the WAPID technology to damp oscillations when updating the Lagrange multiplier.
\end{itemize}

\section{Relate Works}

Constrained reinforcement learning (CRL) enforces constraint satisfaction on the expectation of cost function while maximizing the expected return.
In this paper, we mainly discuss the Lagrangian relaxation methods and other related RL methods that consider constraints. 

The Lagrangian relaxation method is widely applied to CRL. For example, \cite{chow2017risk} employs the Lagrangian method to devise policy gradient and actor-critic algorithm for risk-constrained RL. 
\cite{tessler2018reward} proposes RCPO and proves this approach converge to a saddle point under their described assumptions. 
\cite{Ray2019} designs a safety exploration environment, named safety gym.
To obtain TRPO-Lagrangian and PPO-Lagrangian, \cite{Ray2019} combines the Lagrangian approach with trust-region policy optimization (TRPO) \cite{schulman2015trust} and proximal policy optimization (PPO) \cite{schulman2017proximal}. 
\cite{bohez2019value} extends a single constraint to a set of point-wise constraints by introducing state-dependent Lagrange multipliers.
\cite{stooke2020responsive} proposes the PID-Lagrangian method to handle the oscillation problem which affects the performance of Lagrangian approach. \cite{yang2021wcsac} adopts conditional Value-at-Risk to SAC-Lag method and focuses on the upper tail of the cost distribution. 
\cite{ma2021feasible} proposes the feasible actor-critic (FAC) algorithm to guarantee state-wise safety with a multiplier network as the feasibility indicator. 
Similar to the above approaches, our method is also a Lagrangian-based method. Lagrangian method is simple to implement and easy to combine with other methods. Furthermore, our method makes the agent's exploration process safer in a conservative manner compared to the above methods.

Meanwhile, some works do not utilize Lagrangian relaxation to solve the CRL problem. 
For example, \cite{achiam2017constrained} proposes CPO to handle the constraints by approximating the constrained optimization problem with a quadratic constrained optimization problem.
In addition, \cite{neely2010stochastic,chow2019lyapunov} adopt Lyapunov functions to solve CRL. 
Specifically, \cite{chow2019lyapunov} develops $\theta$-projection and $\alpha$-projection, which are two classes of policy optimization algorithms based on Lyapunov functions, to learn safe policies. 
Based on CPO, PCPO \cite{yang2020projection} utilizes TRPO to update the policy with a quadratic approximation. Note that PCPO provides a way to recover from an infeasible set and presents theoretical performance bounds. 
However, because these approaches update policies with the quadratic approximation, they also have the drawbacks of expensive computation and limited generality.

\section{Preliminaries}
\subsection{Constrained Markov Decision Processes}
In this paper, we consider the safety constraint optimization problem as a constrained Markov decision process (CMPD). 
A typical CMDP consists of state $s\in \cal{S}$, action $a\in \cal{A}$, reward $r(s,a)\in \mathbb{R}$, cost $c(s,a) \in \mathbb{R}$, transition probabilities $p(s^{\prime}|s,a)$, a discount factor $\gamma \in [0,1)$, and a given safety threshold $d$. 
$S$ and $A$ correspond to the state set and the action set, respectively. The reward function $r(s,a)$ indicates the instant reward after taking action $a$ in state $s$. The cost function $c(s,a)$ evaluates whether the constraints are satisfied under the current $(s, a)$. The transition function $p(s^{\prime}|s,a)$ defines the probability of changing from state $s$ to $s^{\prime}$ when taking action $a$. 
We define the estimation of the expected discounted return for a given policy as the Q-function:
\begin{equation}
    Q(s,a) = \mathbb{E} [\sum_t \gamma^t r(s_t,a_t)|s_0=s,a_0=a].
\end{equation}
Similarly, the C-function is defined as:
\begin{equation}
    C(s,a) = \mathbb{E} [\sum_t \gamma^t c(s_t,a_t)|s_0=s,a_0=a].
\end{equation}

\subsection{Lagrangian Relaxation}
The goal of CMDP is to maximize the expected return of the task, while ensuring that the expected cost satisfies the constraints.
The objective function is described as:
\begin{equation}
\begin{aligned}
&\max_{\pi }\mathbb{E}_{( s,a) \sim \rho _{\pi }}\left[\sum _{t} \gamma ^{t} r( s,a)\right], \\
&\text{s.t.}\ \mathbb{E}_{( s,a) \sim \rho _{\pi }}\left[\sum _{t} \gamma ^{t} c( s,a)\right] \leq d.
\end{aligned}
\label{equ_objective}
\end{equation}
This constraint problem can usually be solved using the Lagrange-multiplier method. We use Lagrangian relaxation to transform Eq. \eqref{equ_objective} into the corresponding unconstrained dual problem as follows:
\begin{equation}
    \min_{\lambda \geq 0} \max_{\theta } L( \theta ,\lambda ) =J_{R}^{\pi _{\theta }} -\lambda \left( J_{C}^{\pi _{\theta }} -d\right),
\end{equation}
where
\begin{equation}
    J_{R}^{\pi _{\theta }}=\mathbb{E}_{( s,a) \sim \rho _{\pi }}\left[\sum _{t} \gamma ^{t} r( s,a)\right],
\end{equation}
and 
\begin{equation}
    J_{C}^{\pi _{\theta }}=\mathbb{E}_{( s,a) \sim \rho _{\pi }}\left[\sum _{t} \gamma ^{t} c( s,a)\right].
\end{equation}

\section{Method}
\subsection{Overview}
Based on Maximum a posteriori Policy Optimisation (MPO) \cite{abdolmaleki2018maximum}, we propose the Conservative Distributional MPO (CDMPO) algorithm to deal with safety constraints.
CDMPO introduces safety constraints into the optimization objective by leveraging the Lagrangian function.
We expect the agent to adopt a more conservative strategy during training to reduce the overall cost.
To this end, we introduce a conservative exploration approach to select the most conservative action from the sampling action set. 
Meanwhile, we design a more conservative objective function for the C-function.
In addition, we introduce safety constraints based on the policy improvement of MPO, and derive a new loss function for policy learning under safety constraints.
Finally,  we introduce the Weighted Average Proportional-Integral-Derivative (WAPID) technique to stabilize the optimization of the Lagrange multiplier term.

\subsection{Conservative Distributional C-function}
In this subsection, we describe how to construct a conservative exploration method and how to conservatively conduct policy evaluation.

In order to ensure the safety of the agent when exploring the environment, we use the C-function to assist in selecting the actions. Specifically, the actor network outputs an action set $A(s,n)$ according to the current state $s$ and parameter $n$. $n$ represents the number of actions. Then, C-network takes state $s$ and the action set $A(s,n)$ as input. Finally, the proposed method selects the action with the smallest $C(s, a)$. The safety exploration method is formalized as:
\begin{equation}
    a = \argmin_{a^\prime \in A(s,n)} C(s,a^\prime).
\label{epu7}
\end{equation}

As the choice of action depends on the action set $A(s,n)$ and the C-function, we consider how to learn an approximate Q-function and an approximate C-function with credibility. Inspired by value distribution methods \cite{bellemare2017distributional}, we treat long-term returns as a random variable $Z_R^{\pi}$, and long-term costs as a random variable $Z_C^{\pi}$. The value function of them are as follows:
\begin{gather}
    Q_\pi(s,a)=\mathbb{E}Z_R^{\pi},\\ 
    C_\pi(s,a)=\mathbb{E}Z_C^{\pi}.
\end{gather}

We utilize discrete distributions to represent distributions of above two random variables.
There is a hyperparameter $N \in \mathbb{N}$, which discretizes the distribution of value into N segments. We formulate discrete distributions as $\{z_i=V_{min}+i \triangle z : 0\leq i<N\}$, $\triangle z=\frac{V_{max}-V_{min}}{N-1}$, where $V_{min}$ and $V_{max}$ represent the estimated range of random variables.
The probabilities of each segment are given by a parametric model $\theta$, which is a linear layer from the state-action to the logits, followed by a softmax activation. 
\begin{equation}
    Z_{\theta}(s,a) = z_i, \quad \text{w.p.} \quad   p_i(s,a)=\frac{e^{\theta_i(s,a)}}{\sum_je^{\theta_j(s,a)}}.
\end{equation}

The distributional Bellman operator can be defined as:
\begin{gather}
  (\mathcal{T}Z_R^{\pi_\theta})(s,a)=r(s,a)+\gamma \mathbb{E}[Z(s^\prime,\pi(s^\prime))|s,a], \\
  (\mathcal{T}Z_C^{\pi_\theta})(s,a)=c(s,a)+\gamma \mathbb{E}[Z(s^\prime,\pi(s^\prime))|s,a].
\end{gather}
Their Temporal Difference (TD) losses can be written as the cross-entropy term of the KL divergence and formalized as:
\begin{equation}
    L_{TD}(\theta)=\mathbb{E}_{(s,a)}[\sum_{i=0}^{N}p^\prime_i\frac{e^{\theta_i(s,a)}}{\sum_je^{\theta_j(s,a)}}].
\label{epu_TD}
\end{equation}
Note that $p^\prime$ denote the probabilities of the projected distributional Bellman operator \cite{bellemare2017distributional}, and the operator is applied to some target distribution $Z_{target}$. 

Additionally, we also expect that the evaluation of the C-function is conservative. To train such a conservative critic $C(s,a)$, we draw lessons from CQL Loss \cite{kumar2020conservative} and construct a new loss to estimate $C(s,a)$. The Conservative Distributional C-function Learning (CDCL) loss is:
\begin{equation}
    \begin{aligned}
        L_C(\theta) =& \beta(\mathbb{E}_{(s,a) \sim \mathcal{\beta}}[C(s,a)] - \mathbb{E}_{s \sim \mathcal{\beta}, a \sim \pi(s)}[C(s,a)])  \\
        & + L_{TD}(\theta).
    \end{aligned}
\label{epu_CDCL}
\end{equation}
Note that $\beta$ is to control the relative importance of the regularization term and the TD loss.
$\mathbb{E}_{(s,a) \sim \mathcal{\beta}}[C(s,a)]$ is to minimize the C-function evaluation of the corresponding state-action pair in the replay buffer. Note that only the conservative actions are stored in the replay buffer. 
$\mathbb{E}_{s \sim \mathcal{\beta}, a \sim \pi(s)}[C(s,a)])$ is to maximize the expectation of C-function over actions sampled from the current policy. 
$L_{TD}(\theta)$ is the distributional TD loss of the C-function. 

\subsection{Policy Improvement with Safety Constraints}
In this part, we describe how to improve the policy under safety constraints.
We directly introduce safety constraints into the objective function of general policy improvement, and maximize it. The new objective function $J(\theta)$ is:
\begin{equation}
    J( \theta) =E_{( s,a) \sim \rho _{\pi }}[ Q_{\pi _{\theta }}( s,a) -\lambda ( C_{\pi _{\theta }}( s,a) -d)],
\end{equation}
where $\lambda$ is treated as a constant. In practice, we execute policy evaluation using MPO, which can be divided into E-step and M-step.

\subsubsection{E-step}
In the E-step, MPO calculate following KL regularized RL objective as follows:
\begin{equation}
    \begin{split}
     &\max_{q}\int \mu ( s)\int q( a|s)[ Q( s,a) -\lambda ( C( s,a) -d)] \mathrm{d}a\mathrm{d}s\\
     &\text{s.t.}\quad  \left\{\begin{array}{lc}
     \ \int \mu ( s) \text{KL}( q( a|s) ,\pi ( a|s,\theta _{t})) \mathrm{d}a\mathrm{d}s < \epsilon \\
     \ \iint \mu ( s) q( a|s) \mathrm{d}a\mathrm{d}s=1
     \end{array}\right.
    \end{split}
    \label{equ_16}
\end{equation}

We can derive an approximate solution (see the appendix for a full derivation) as:
\begin{equation}
    \begin{aligned}
         q( a|s) =&\pi ( a|s,\theta _{i}) \exp\left(\frac{Q_{}( a,s)}{\eta }\right) \exp\left(- \frac{\eta -\gamma }{\eta }\right) \\ 
         &\cdot \exp\left( -\frac{\lambda ( C_{}( a,s) -d)}{\eta }\right).
    \end{aligned}
    \label{equ_qsa}
\end{equation}
Then we can obtain $\eta$ by minimizing the following convex dual function:
\begin{equation}
    \begin{aligned}
      g( \eta ) =&\eta \int \mu _{q}( s) log(\int \pi ( a|s,\theta _{i}) \exp(\frac{Q( a,s)}{\eta }) \\
      &\cdot\exp( -\frac{\lambda ( C_{}( a,s) -d)}{\eta }) \mathrm{d}a)\mathrm{d}s + \eta \epsilon.
    \end{aligned}
    \label{equ_eta}
\end{equation}

\subsubsection{M-step}
After the E-step, we get an improved sample-based distribution over actions for each state in the non-parametric case. 
Then, we generalize this sample-based solution $q(a|s)$ over state and action space. 
The above process can be formulated as the following weighted supervised learning problem:
\begin{equation}
    \pi_{\theta_{k+1}}=\argmax_{\pi_\theta}\mathbb{E}_{(s,a)}[q(a|s)\log\pi_\theta(a|s)],
    \label{equ_mstep_wsl}
\end{equation}
where $\theta$ are the parameters of our function approximator. 
We use the weights of the previous policy $\pi_{\theta_k}$ to initialize $\theta$.

In order to reduce the overfitting of the sample distribution, we introduce the KL divergence constraint  to limit the overall change in the parametric policy \cite{abdolmaleki2018maximum,abdolmaleki2018relative} as follows:
\begin{equation}
\mathbb{E}_{\mu_q(s)}[\text{KL}(\pi_{\theta_k}(a|s), \pi_{\theta}(a|s))]<\epsilon.
\label{equ_mstep_kl}
\end{equation}

\begin{algorithm}[tb]
\caption{Weighted Average PID Lagrange Multiplier}
\label{alg:algorithm_pid}
\textbf{Input}: $J_C$\\
\textbf{Parameter}: $K_P,K_I,K_D\ge 0$, $0 \leq w \leq 1$\\
\textbf{Output}: $\lambda$
\begin{algorithmic}[1] 
\STATE Integral: $I\leftarrow 0$.
\STATE Previous Cost: $J_{C,prev} \leftarrow 0$.
\FOR{each iteration}
    \STATE Receive cost $J_C$
    \STATE $\vartriangle \leftarrow J_C-d$
    \STATE $\partial \leftarrow (J_C-J_{C,prev})_+$
    \STATE $I \leftarrow I + w(\vartriangle - I)_+$
    \STATE $\lambda \leftarrow (K_P\vartriangle+K_II+K_D\partial)_+$
    \STATE $J_{C,prev} \leftarrow J_C$
    \STATE \textbf{return} $\lambda$
\ENDFOR

\end{algorithmic}
\end{algorithm}

\subsection{Weighted Average PID Lagrange Multiplier}
In Eq. (\ref{equ_16}), it is difficult to stably optimize the Lagrange multiplier $\lambda$ by the gradient descent method.
\cite{stooke2020responsive} analyzes that optimizing Lagrange multiplier exhibits oscillations and overshoots that destabilize the policy learning. 
Therefore, we adopt a Proportional Integral Derivative (PID) controller to update $\lambda$ with fast and steady learning process. The update procedure is formulated as:
\begin{equation}
    \lambda \leftarrow K_p\delta + K_i\int_{i=1}^{k}\delta di + K_d \frac{\delta}{d_i},
\end{equation}
where $\delta=\mathbb{E}_{(s,a)}[C(s,a)-d]$.

However, if the constraint is very tight and the
initial policy is relatively unsafe, the integral terms usually
increase rapidly since $\delta$ is large. The original integral term $int$ update formula is as follows:
\begin{equation}
int_{k+1} = int_{k} + \delta.
\label{equ23}
\end{equation}

We introduce the weight parameter $w$ to rewrite Eq. (\ref{equ23}) as follows:
\begin{equation}
int_{k+1} = int_{k} + w(\delta - int_{k}),
\end{equation}
where the parameter $w \in (0, 1]$ is a constant. This results in $int_{k+1}$ being a weighted average of past $\delta$ and the initial estimate $int_1$:
\begin{equation}
    \begin{aligned}
      int_{k+1} &= int_{k} + w(\delta_k - int_{k}) \\
      &=w\delta_k + (1-w)int_{k} \\
                     &= (1-w)^n int_1 + \sum_{i=1}^{k}w(1-w)^{n-i} \delta_i.
    \end{aligned}
\end{equation}
Algorithm \ref{alg:algorithm_pid} is our Weighted Average PID Lagrange Multiplier (WAPID) method.
Algorithm \ref{alg:algorithm_cdmpo} is our complete approach, combining conservative distributional C-function, policy improvement with safety constraints, and WAPID. 

\begin{algorithm}[tb]
\caption{CDMPO}
\label{alg:algorithm_cdmpo}
\textbf{Parameter}: $n \ge 1$, policy $\theta$, Q-function $Q(s,a)$, C-function $C(s,a)$
\begin{algorithmic}[1] 
\FOR{each iteration}
\FOR{each environment step}
\STATE $A(s_t,n) = {a_1,a_2, \cdots, a_n}\sim \pi_\theta(s_t)$
\STATE $a_t = \argmin_{a^\prime in A(s,n)} C(s_t,a^\prime)$
\STATE $s_{t+1} \sim p(s_{t+1}|s_t,a_t)$
\STATE $D \leftarrow D \cup \{(s_t,a_t,r(s_t,a_t),c(s_t,a_t),s_{t+1})\}$
\ENDFOR
\FOR{each gradient step}
\STATE Sample experience from replay buffer $D$
\STATE //policy evaluation
\STATE update Q-fnction according to Eq. \eqref{epu_TD} 
\STATE update C-fnction according to Eq. \eqref{epu_CDCL}
\STATE //policy improvement, update $\pi_\theta$
\STATE calculate $q(s,a)$ according to Eq. \eqref{equ_qsa} and Eq.\eqref{equ_eta}
\STATE update $\theta$ according to Eq. \eqref{equ_mstep_wsl}, \eqref{equ_mstep_kl} and $q(s,a)$ 
\STATE //Lagrangian update
\STATE update $\lambda$ according to algorithm \ref{alg:algorithm_pid} 
\ENDFOR
\ENDFOR

\end{algorithmic}
\end{algorithm}

\begin{figure*}[!t]
\centering
\subfloat[CarGoal1 CDMPO VS on-policy]{\includegraphics[width=\columnwidth]{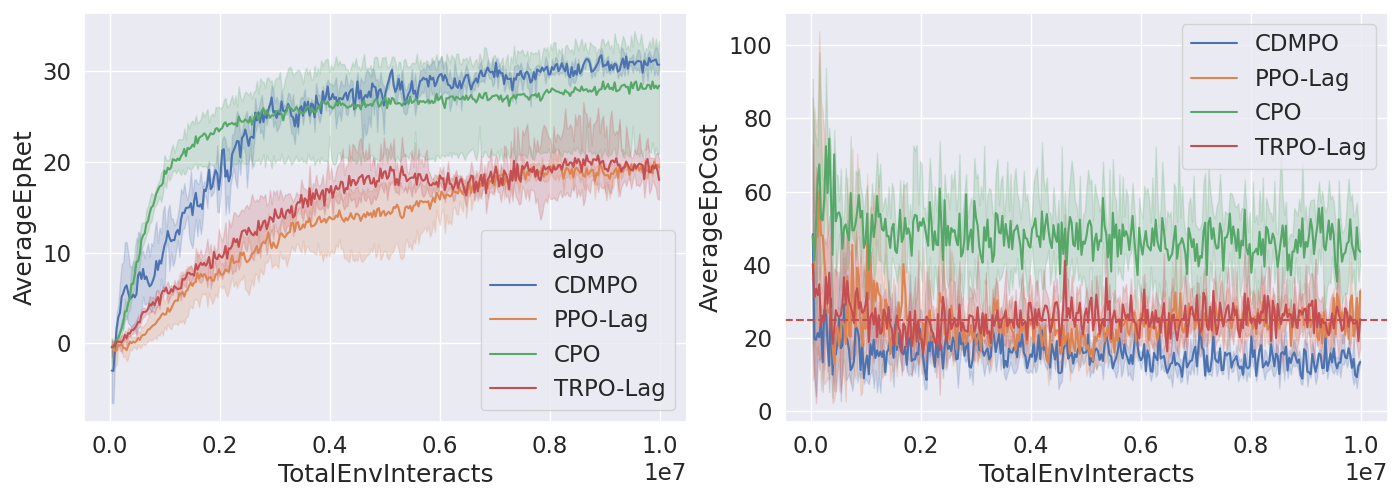}
\label{fig:cdmpo_onpolicy_offpolicy_a}}
\hfil
\subfloat[CarGoal1 CDMPO VS off-policy]{\includegraphics[width=\columnwidth]{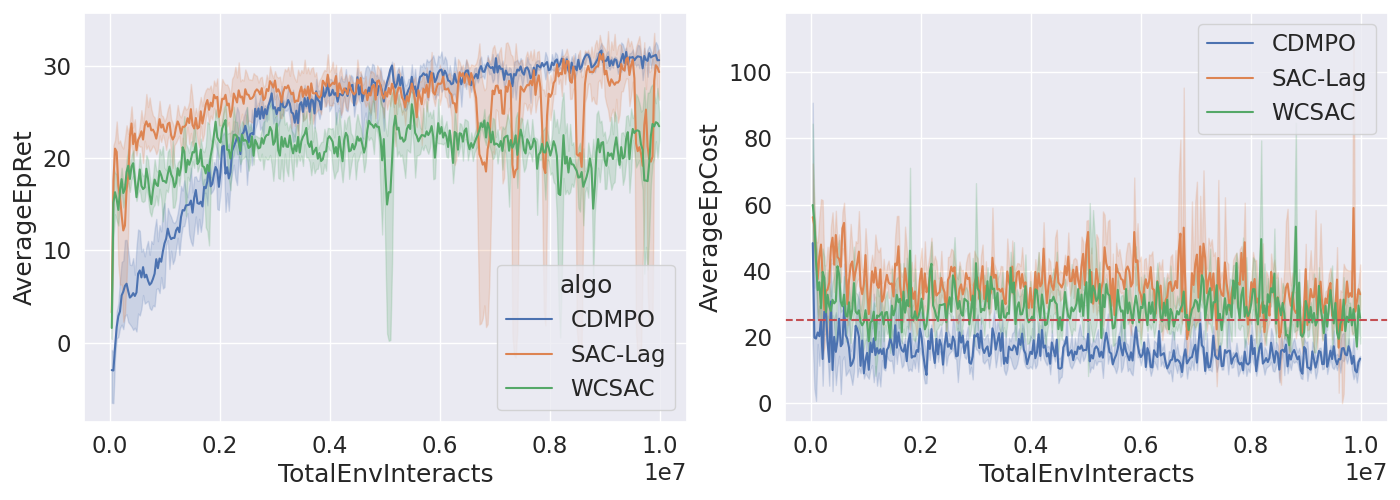}
\label{fig:cdmpo_onpolicy_offpolicy_b}}
\hfil
\subfloat[PointGoal1 CDMPO VS on-policy]{\includegraphics[width=\columnwidth]{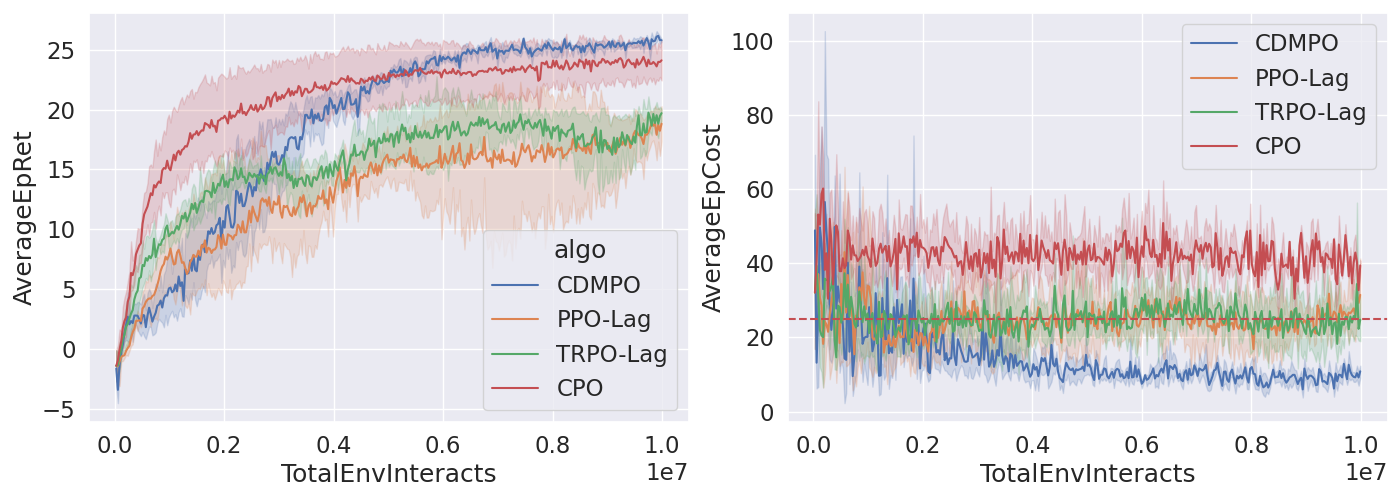}
\label{fig:cdmpo_onpolicy_offpolicy_c}}
\hfil
\subfloat[PointGoal1 CDMPO VS off-policy]{\includegraphics[width=\columnwidth]{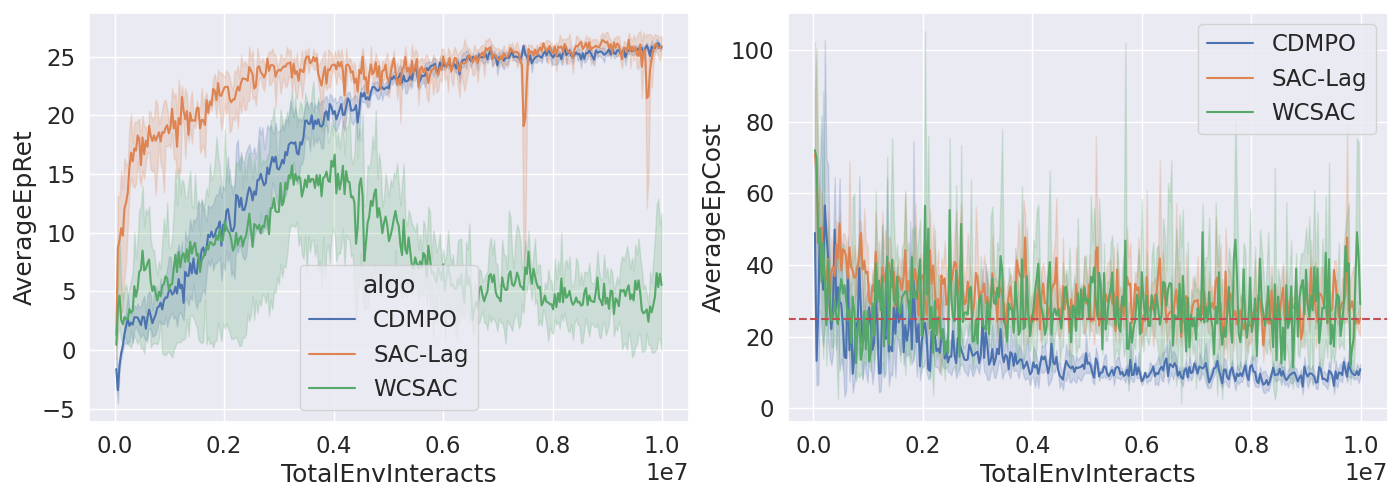}
\label{fig:cdmpo_onpolicy_offpolicy_d}}

\caption{Comparison of PPO-Lag, TRPO-Lag, CPO, SAC-Lag, WCSAC and CDMPO during training in CarGoal1(top row) and PointGoal1(botton row) . The lines are the average of four runs, and the shaded area is the standard deviation.}
\label{fig:cdmpo_onpolicy_offpolicy}
\end{figure*}

\section{Experiments}
We conduct experiments in the safety-gym \cite{Ray2019} environment. Safety-gym is a robot control environment and focuses on safety exploration. 
We evaluate the algorithms on the CarGoal1 and PointGoal1 tasks that involve steering the robot to a series of goal locations. 

Each task has multiple different types of hazards and pillars, which induce a cost when contacted by the robot (without necessarily hindering its movement). Hazards are placed randomly at each episode and often lay in the path to the goal.
The robot senses the position of hazards and the goal via a coarse and LIDAR-like mode. The reward of the environment consists of two parts. The first part is the density reward, which encourages the robot moving toward the goal. The second part is a large and sparse reward when the robot achieves the goal. Additionally, the location of the robot and the goal will be reset randomly when robot reach a goal during training.

\subsection{Comparison Methods}
\begin{itemize}
    \item Constrained Policy Optimization (CPO): \cite{achiam2017constrained} enforces constraints throughout training with guarantees for near-constraint satisfaction at each iteration based on theoretical analysis.
    \item On-policy Lagrangian methods: TRPO-Lag and PPO-Lag \cite{Ray2019} combine the Lagrangian approach with TRPO and PPO using adaptive penalty coefficients to enforce constraints
    \item Off-policy Lagrangian methods: SAC-Lag \cite{ha2020learning} combines SAC with Lagrangian method leads to a safety-constrained RL framework to address local constraints (constraints are set for each timestep instead of each episode). WCSAC \cite{yang2021wcsac} is based on SAC-Lag and adopt conditional Value-at-Risk to focus on the upper tail of the cost distribution.
\end{itemize}

\subsection{Implementation Details}
We implement CDMPO based on Acme \cite{hoffman2020acme} codebase. Related parameters are illustrated in appendix. Our experiments are performed with Intel(R) Xeon(R) CPU E5-2620 v4 and a NVIDIA 2080 GPU.
We set $d=25$ as cost limit for CarGoal1 and PointGoal1 task. All the methods are trained for 10 million steps. We record the evaluation metrics of algorithms for each 30,000 steps. 

\begin{figure}[!t]
\centering
\subfloat[Test performance in CarGoal1]{\includegraphics[width=0.99\columnwidth]{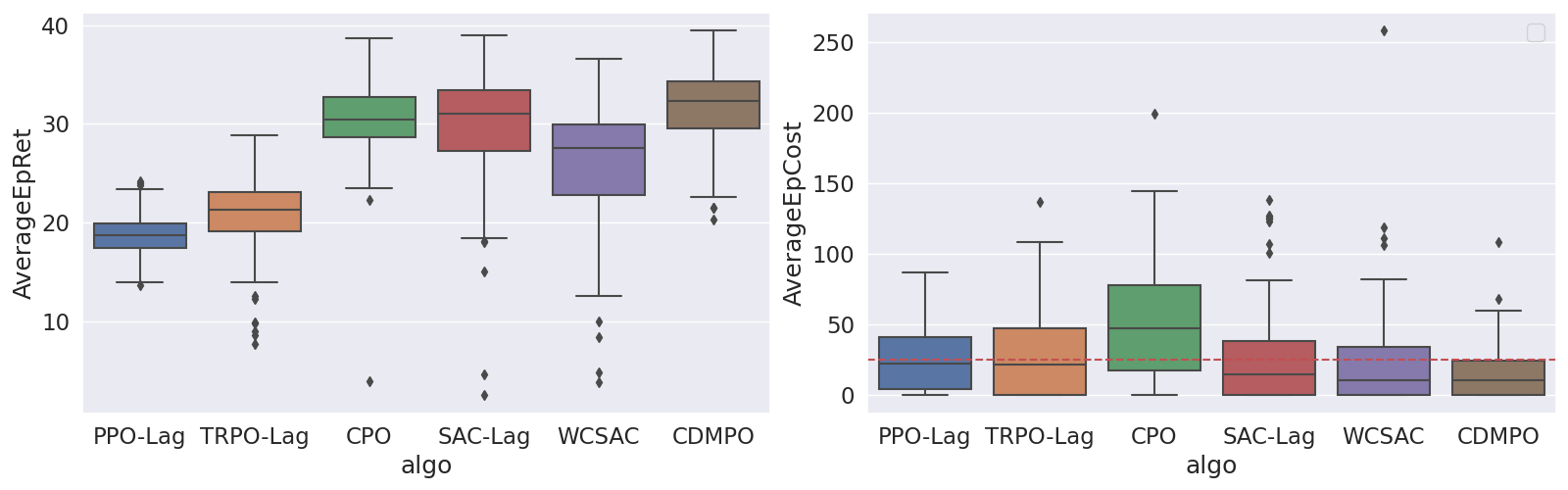}
\label{fig:boxplot_cdmpo_a}}
\hfil
\subfloat[Test performance in PointGoal1]{\includegraphics[width=0.99\columnwidth]{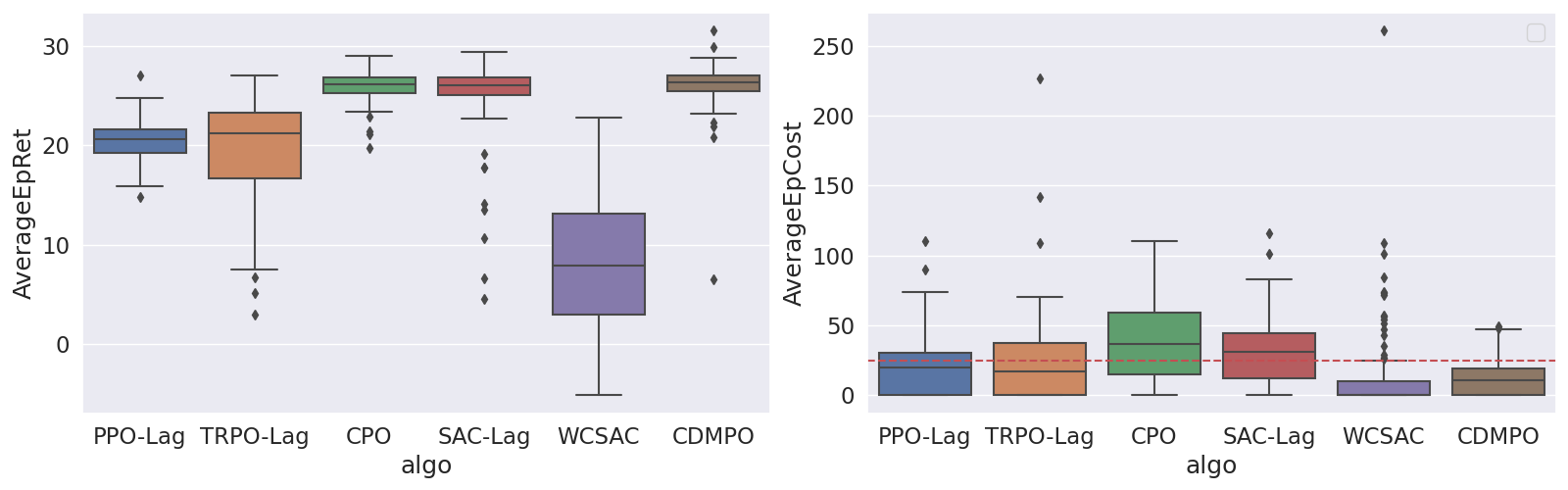}
\label{fig:boxplot_cdmpo_b}}
\caption{The test performance for the final policies of each algorithm. Every algorithm runs for 100 episodes. Environment: CarGoal1 (top row), PointGoal1 (botton row), cost limit 25.}
\label{fig:boxplot_cdmpo}
\end{figure}

\subsection{Comparison During Training}
In this subsection, we focus on the performance of the algorithm during the training, including average episodic return and average episodic cost. Please refer to Figure \ref{fig:cdmpo_onpolicy_offpolicy} for details.
For a clearer comparison of algorithms, we compare the on-policy algorithm (PPO-Lag, TRPO-Lag, CPO) in the Fig \ref{fig:cdmpo_onpolicy_offpolicy_a}, Fig \ref{fig:cdmpo_onpolicy_offpolicy_c} and the off-polciy algorithm (SAC-Lag, WCSAC) in the Fig \ref{fig:cdmpo_onpolicy_offpolicy_b} and Fig \ref{fig:cdmpo_onpolicy_offpolicy_d}. 

Figure \ref{fig:cdmpo_onpolicy_offpolicy} shows that CDMPO achieves the best performance on the episodic return and generates fewer constraint violations during training.
CPO fails to satisfy the expectation-based constraints in two safety exploration tasks, which is consistent with observation made by \cite{Ray2019}.
SAC-Lag has a good performance on task rewards but often violates constraints.
Although WCSAC is competitive with CDMPO considering the behavior of violating constraints, it sacrifices the performance of task rewards a lot.
PPO-Lag and TRPO-Lag perform well to satisfy the expectation-based constraints, but they are not competitive in term of task rewards.

\subsection{Evaluation During Testing}

In this subsection, we focus on the final performance achieved by the trained algorithms.
After training, we use 100 episodes to evaluate the trained policies of each algorithm. The results can be seen in Figure \ref{fig:boxplot_cdmpo}. For PPO-Lag, TRPO-Lag and SAC-Lag, although they yield a constraint-satisfying result for their expectation-based constraints (their median line is below the constraint threshold), there are still half of the episodes suffer from dangerous. It suggests that expectation-based safety constraints are not enough to guarantee the safety of a trajectory.
In addition, WCSAC reduces the number of constraint violations by significantly sacrificing the performance of the task reward. Contrary to the above algorithms, CDMPO  has competitive performance in the task reward. Meanwhile, the most episodes of CDMPO satisfy the constraints.

\begin{figure}[t]
\centering
\includegraphics[width=\columnwidth]{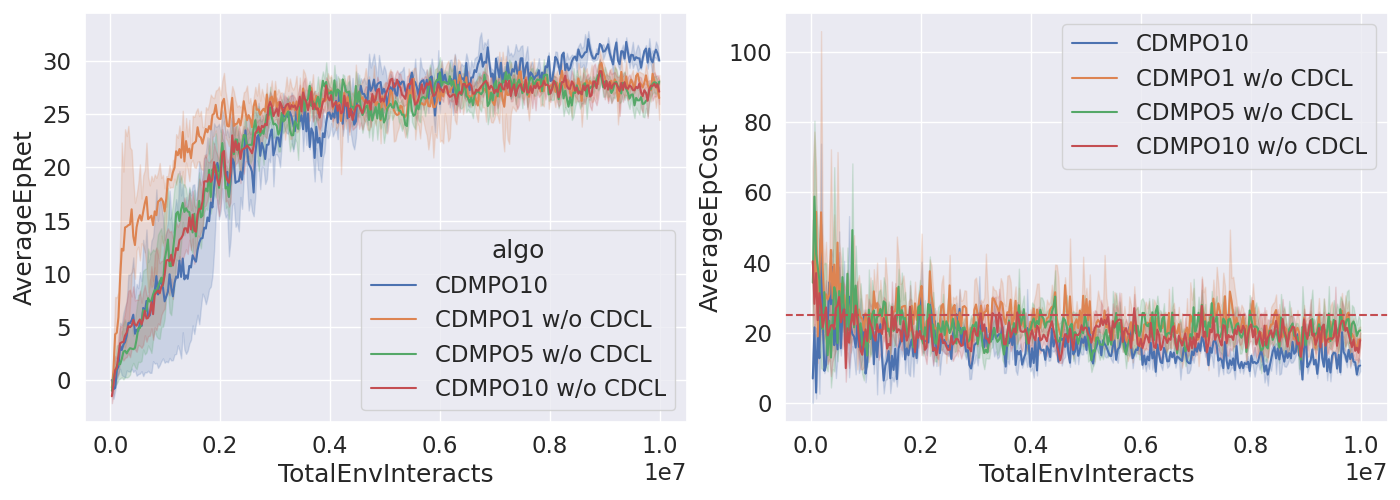} 
\caption{Different ablation performances of CDMPO during training. ``w/o CDCL'' means do not use CDCL loss. Environment: CarGoal1, cost limit 25}.
\label{fig:cdmpo_1510}
\end{figure}

\begin{table}[t]
	
	\centering
	\small
	\resizebox{1 \linewidth}{!}{
	\begin{tabular}{cc}
		\toprule
		Algorithm & Number of Constraint Violations    \\ 
		\midrule
	    CDMPO1 w/o CDCL & 109 \\
		CDMPO5 w/o CDCL & 74 \\
		CDMPO10 w/o CDCL & 60 \\
		CDMPO10 & 20\\

		\bottomrule
	\end{tabular}
	}
	\caption{The number of constraint Violations by different algorithms during training.}
	\label{tab:cdmpo1510-test}
	\vspace{-10pt}
\end{table}

\begin{figure}[t]
\centering
\includegraphics[width=\columnwidth]{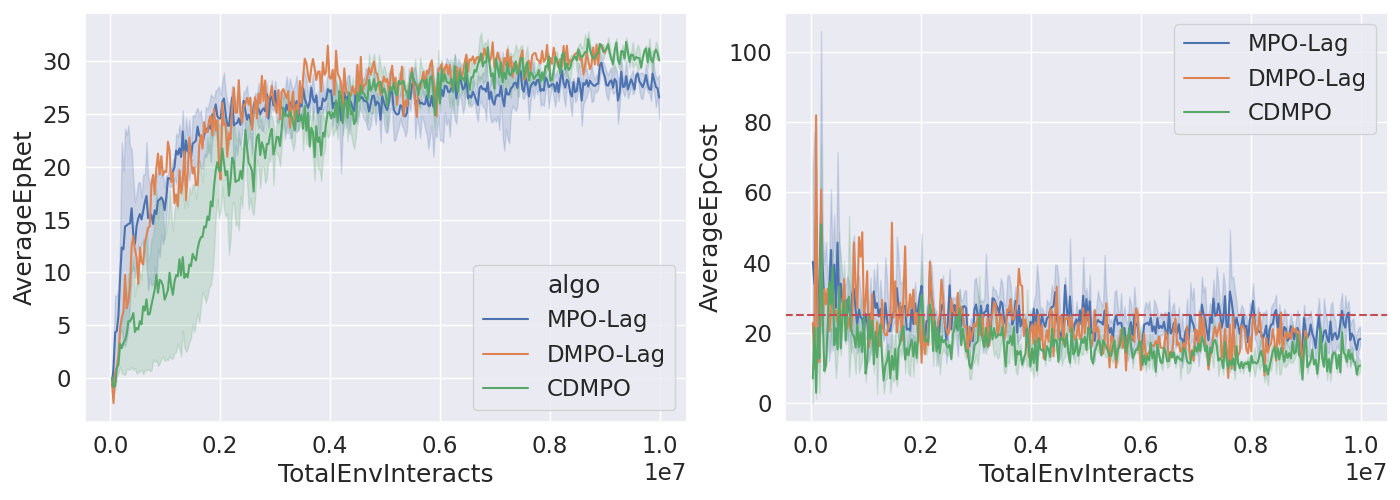} 
\caption{Performance of different algorithms during training, which corresponds to section 5.5(2). Environment: CarGoal1, cost limit 25}.
\label{fig:mpo_dmpo_cdmpo}
\end{figure}

\subsection{Ablation Study}
In this subsection, we discuss the impact of the proposed components. Specifically, We discuss the following issues: 


(1) \textit{Whether the conservative exploration approach can avoid risks, and what dose the number of elements in the action candidate set affect?}
We count the number of violations of constraints during training and list in Table \ref{tab:cdmpo1510-test}. 
Figure \ref{fig:cdmpo_1510} and Table \ref{tab:cdmpo1510-test} show that the number of violations is inversely proportional to the number of actions in the candidate set.

(2) \textit{Whether utilizing the discrete value distribution to estimate Q-function and C-function will bring benefits?}
In Figure \ref{fig:mpo_dmpo_cdmpo}, we compare DMPO-Lag and MPO-Lag. 
MPO-Lag is a basic MPO method that adds Lagrangian safety constraints which described in Section 4.3, and utilizes WAPID techniques. 
DMPO-Lag uses distributed reinforcement learning to estimate Q-function and C-function based on the former algorithm. The results show that DMPO-Lag has better performance and fewer violations than MPO-Lag.

(3) \textit{Does adopting CDCL loss contribute to the stable update of C-function?}
To demonstrate the positive impact of CDCL loss, we compare CDMPO10 w/o CDCL and CDMPO10, and the difference between them is whether CDCL loss is used during training. Figure \ref{fig:cdmpo_1510} and Table \ref{tab:cdmpo1510-test} show that CDMPO10 has better performance on the task reward while having fewer constraint violations. This shows that CDCL loss has a positive effect on the training of the algorithm.

\begin{figure}[t]
\centering
\includegraphics[width=\columnwidth]{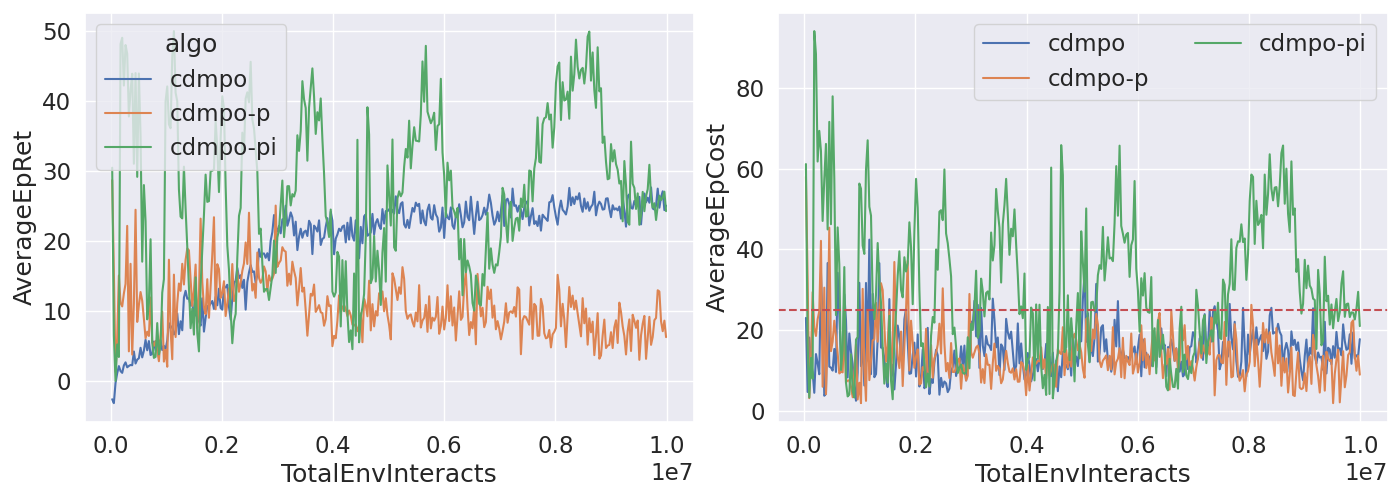} 
\caption{Different types of PID control on the Lagrange multiplier damps oscillations. Blue line cdmpo (means use WAPID), orange line cdmpo-p (means only use proportional control)), green line cdmpo-pi(means use PID). Environment: CarGoal1, cost limit 25}.
\label{fig:cdmpo_pid}
\end{figure}

(4) \textit{Whether the WAPID Lagrange Multiplier can stabilize the optimization of the Lagrange multiplier?}
We compare the effect of the WAPID and PID methods in Figure \ref{fig:cdmpo_pid}. The orange line (only use proportional control) can not achieve good results in the task reward because there exists steady-state error at convergence, while the green line (PID methods) frequently violates constraints and the task reward performance is highly erratic. In contrast, the blue line (WAPID) can satisfy the constraints while having good performance. Above results prove that WAPID can stabilize the optimization of the Lagrange multiplier.


\section{Conclusion}
In this paper, we propose the CDMPO algorithm to solve safety-constrained RL problems. Our method incorporates a conservative exploration strategy as well as a conservative distribution function. This enables the agent to explore safely during training. In addition, we use the WAPID technique to make the training process more stable. 

In this paper, the distribution of long-term costs is approximated to be a discrete distribution. In the future, we can further explore modeling more complex distribution. Then we will solve the safety problems in  practical autonomous driving scenarios.

\bibliographystyle{named}
\bibliography{ijcai22}

\end{document}